# On Regularization Properties of Artificial Datasets for Deep Learning

K. ANTCZAK

karol.antczak@wat.edu.pl

Military University of Technology, Faculty of Cybernetics
Institute of Computer and Information Systems
Kaliskiego Str. 2, 00-908 Warsaw, Poland

The paper discusses regularization properties of artificial data for deep learning. Artificial datasets allow to train neural networks in the case of a real data shortage. It is demonstrated that the artificial data generation process, described as injecting noise to high-level features, bears several similarities to existing regularization methods for deep neural networks. One can treat this property of artificial data as a kind of "deep" regularization. It is thus possible to regularize hidden layers of the network by generating the training data in a certain way.

## 1. Generalization gap

A distinguishing feature of machine learning models is an ability to work on previously unseen data. Such ability, known as *generalization* [1], can be formally expressed by means of *generalization gap*, defined as a discrepancy between mean losses for the training dataset $X_{train}$ and the whole dataset $X$ for some model $\theta$:

$$G(\theta, X_{train}, X) = L(\theta, X) - L(\theta, X_{train}) \quad (1)$$

Generalization plays a crucial role in machine learning since, in real-life applications, training dataset is a small subset of all possible examples from a problem domain, i.e. $X_{train} \subseteq X$. In the case of a labeled dataset, it is possible to create a trivial model that has zero error on the training dataset, simply by assigning proper label to known sample. Despite this, it will not be very useful. Since:

$$L(\theta, X) = \frac{1}{n}\sum_{i=0}^{n} L(\theta, x_i),$$
$$X = \{x_0 \dots x_n\} \quad (2)$$

The generalization gap will be:

$$G(\theta, X_{train}, X) = \frac{1}{n}\sum_{i=0}^{n} L(\theta, x_i) - \frac{1}{m}\sum_{j=0}^{m} L(\theta, x_{train_j}) =$$
$$= \frac{1}{n}\sum_{i=0}^{n} L(\theta, x_i) \quad (3)$$

In the general case, there is no upper bound for generalization gap of such model. Thanks to the "no free lunch" theorem [2] we know that similar weakness is a common property of all machine learning models: It is not possible to create a general model that will have minimal generalization gap for all types of problems. On the other hand, deep learning use number of assumptions regarding problem space and techniques that allow them to perform exceedingly well in many real-life scenarios.

Note that it is usually not possible to calculate an exact value of the generalization gap simply because the whole dataset $X$ is not known. Instead, it be can estimated with the following formula:

$$G(\theta, X_{train}, X) \approx L(\theta, X_{val}) - L(\theta, X_{train}) \quad (4)$$

where $X_{val}$ is called the *validation set*. Since $X_{val} \subseteq X$, it is easy to see that such estimator is a consistent one.

## 2. Regularization in deep learning

A number of techniques exist specifically for minimization of generalization gap. They are known as *regularization* methods. In deep learning, many regularization methods are based on the concept of model capacity[1] [3]. It is a function of the model structure that can be interpreted as a measure of complexity of problems that are learnable by the model. A model with sufficiently large capacity is prone to memorize all training samples, resulting in a trivial model shown in the previous section. It is known as an *overfitting problem*. On the other hand, a model with small capacity will not be able to learn the problem with certain complexity, causing an underfitting. Therefore, regularization in deep learning is motivated by searching for the best possible (in terms of generalization gap) model capacity.

---

[1] Measuring a capacity of the model is a difficult problem itself. A number of measures were proposed;

A de facto standard of deep models training algorithms is a *stochastic gradient descent* (SGD) and its variations. It itself is an extension of a gradient descent method that minimizes loss function by iteratively following the direction (in a model parameter space) that reduces the loss the fastest, which is an opposite of the loss gradient:

$$\theta \leftarrow \theta - \eta \nabla_\theta L(X, \theta) \quad (5)$$

$\theta$ is a vector of model parameters (network weights and biases), $\eta$ is a learning rate hyperparameter, $\nabla_\theta L$ is a gradient of the loss function in the model parameter space. The stochastic extension of SGD is that the value of gradient is estimated by calculating mean gradient $g$ for a minibatch of examples, sampled uniformly from the training set:

$$\theta \leftarrow \theta - \eta g(X, \theta)$$
$$g = \frac{1}{m'} \nabla_\theta \sum_{j=0}^{m'} L(\theta, x_j), \quad (6)$$
$$\{x_0 \ldots x_{m'}\} = X_{minibatch} \in X_{train}$$

A number of deep learning regularization methods work through modification of the gradient descent by providing an additional component to the loss function [4]:

$$\tilde{L}(\theta, X) = L(\theta, X) + \alpha R(\theta) \quad (7)$$

$\alpha$ is the regularization hyperparameter and $R(\theta)$ is the regularization function. Following function:

---

the two commonly used in deep learning are Vapnik-Chervonenkis (VC) dimension [15] and Rademacher complexity [14]. In most cases, the exact values of these metrics are not known and only upper and lower bounds are provided for specific types of neural networks.

$$R(\theta) = \|\theta\|_1 = \sum_{i=0}^{n} |\theta_i| \quad (8)$$

is known as $L_1$ regularization. Another used function is:

$$R(\theta) = \|\theta\|_2^2 = \sum_{i=0}^{n} \theta_i^2 \quad (9)$$

This one is known as $L_2$ (or Tikhonov) regularization. Despite formulaic similarity, $L_1$ and $L^2$ regularizations work in a different way. Both of them introduce a penalty for models with large weights, with weight size measured using $L_1$ and $L_2$ norms, respectively. $L_1$ regularization causes an occurrence of an additional component in the loss gradient, equal to $\alpha \text{sign}(\theta)$. For a sufficiently large $\alpha$, it will move weights closer to 0. $L_2$ regularization can be viewed as scaling the model parameters along the axes defined by eigenvectors of loss function Hessian. Model regularized this way will have smaller values for weights associated with input features having small covariance with loss function. Compared to $L_2$, $L_1$ results in a more sparse parameters, i.e. having more zeroes. Since this is usually not the desired behavior of network weights, $L_2$ is more commonly used in deep learning practice.

$L_1$ and $L_2$ regularizations were used long before the "deep learning" term was coined. With the advent of deep learning, specific regularization techniques were developed that take into consideration a structure of neural models. Two modern examples are Dropout and DropConnect methods.

Let us first introduce required conventions. Let $\boldsymbol{y}^{(l)}$ denote the output vector of the layer $l$ of the neural network. A function of the $i$-th neuron in the $(l + 1)$-th layer with the weight vector $\boldsymbol{w}_i^{(l+1)}$, the bias value $b_i^{(l+1)}$ and the activation function $f$ is given by the following formula:

$$y_i^{(l+1)} = f(\boldsymbol{w}_i^{(l+1)} \boldsymbol{y}^{(l)} + b_i^{(l+1)}) \quad (10)$$

A main idea of Dropout [5] is to randomly turn off neurons, along with their outputs, during the training. The modified neuron function will look as follows:

$$y_i^{(l+1)} = f\left(\mathbf{w}_i^{(l+1)} \hat{\mathbf{y}}^{(l)} + b_i^{(l+1)}\right)$$
$$\hat{\mathbf{y}}^{(l)} = \mathbf{r}^{(l)} * \mathbf{y}^{(l)} \quad (11)$$
$$r_j^{(l)} \sim \text{Bernoulli}(p)$$

The probability $p$ of the neuron dropout is a hyperparameter that should be specified before the network training. The $*$ symbol denotes the Hadamard product of two vectors. Values of $r_j^{(l)}$ are sampled during each epoch. After the training, all neurons are used again, therefore it is a common practice to scale the weight vector by a factor of $\frac{1}{p}$ in the trained network.

DropConnect [6] is a newer technique, similar to Dropout in that it drops structural components of the network during the training. In this case, the dropped components are weights rather than neurons. In the vector notation it can be written as:

$$\mathbf{y}^{(l+1)} = f((\mathbf{R}^{(l+1)} * \mathbf{W}^{(l+1)})\mathbf{y}^{(l)})$$
$$r_{ij} \sim \text{Bernoulli}(p) \quad (12)$$

$\mathbf{W}$ denotes the weight matrix and $\mathbf{R}$ is the random matrix with elements $r_{ij}$. Note that for the sake of simplicity the bias vector is included in $\mathbf{W}$.

DropConnect can be viewed as a generalization of Dropout. Assuming that the bias component is included in the $\mathbf{W}$, one can rewrite the Dropout-regularized neuron function in the as a special case of (12), with $\mathbf{R}$ being a diagonal matrix:

$$\mathbf{y}^{(l+1)} = f((\mathbf{R}^{(l+1)} * \mathbf{W}^{(l+1)})\mathbf{y}^{(l)})$$
$$\mathbf{R}^{(l+1)} = \begin{bmatrix} r_1^{(l+1)} & \cdots & 0 \\ \vdots & \ddots & \vdots \\ 0 & \cdots & r_m^{(l+1)} \end{bmatrix} \quad (13)$$
$$r_j^{(l+1)} \sim \text{Bernoulli}(p)$$

Additionally, the $r$ component associated with the bias input will be a constant equal to 1.

Both Dropout and DropConnect work by intentionally damaging the structure of neural network, albeit on a different scale. Their regularization properties can be explained by a similarity to the bagging procedure. Bagging (Bootstrap Aggregating) [7] is a meta-algorithm that reduces variance of model ensemble by producing multiple training sets from a single one by sampling uniformly with replacing. A separate model is then trained for each training set and model ensemble is created, by averaging outputs or voting. Since in Dropout and DropConnect random variables are sampled epoch-wise, it can be viewed as a process of selecting and training a random subnetwork from a larger network during each epoch. For a network consisting of $n$ units, it is possible to sample from $2^n$ possible "thinned" networks. Note, however, that it differs from the classical bagging procedure in that models are not independent from each other and the weights are shared between them. Dropping methods have proven to be effective in practical applications. Combined with their simplicity, it resulted in one of the most popular regularizers of deep neural networks, with many benchmark-winning models using one of them. Nonetheless, Dropout/DropConnect regularizers are not without flaws. Since these methods remove processing units from the structure, they both reduce network capacity. It was shown that the capacity of such regularized networks (measured by means of Rademacher complexity) is a linear function of the probability $p$. As a consequence, the network regularized with Dropout or DropConnect require longer training compared to non-regularized network for the same task.

## 3. Artificial data generation process

Small size of the training dataset is a common reason that prevents deep models from generalization of the acquired knowledge. Two main factors are involved here. The first one is the fact that the smaller dataset has distribution more deviated from the true distribution of the problem space. The second one is that the smaller dataset, the less capacity it requires, making a model prone to overfitting. Therefore, augmentation of the dataset can be used as a technique for reducing the generalization gap. Formally, *data augmentation scheme* [8] is defined as a model for the set $X_{aug}$ created from the set $X_{obs}$ that satisfies the following condition:

$$\int_{\mathcal{M}(X_{aug}) = X_{obs}} p(X_{aug}|y) \mu(dX_{aug}) = p(X_{obs}|y) \quad (14)$$

$\mathcal{M}$ is a mapping of augmented samples to their originals $\mathcal{M}: X_{aug} \to X_{obs}$, $y$ is the output class and $\mu$ is the reference measure on $X_{aug}$. In accordance with the above definition, the marginal distribution of augmented data $p(X_{aug}|y)$ must be the same as the distribution of the original data $p(X_{obs}|y)$.

An internal structure of the training data plays an immense role in deep learning. An

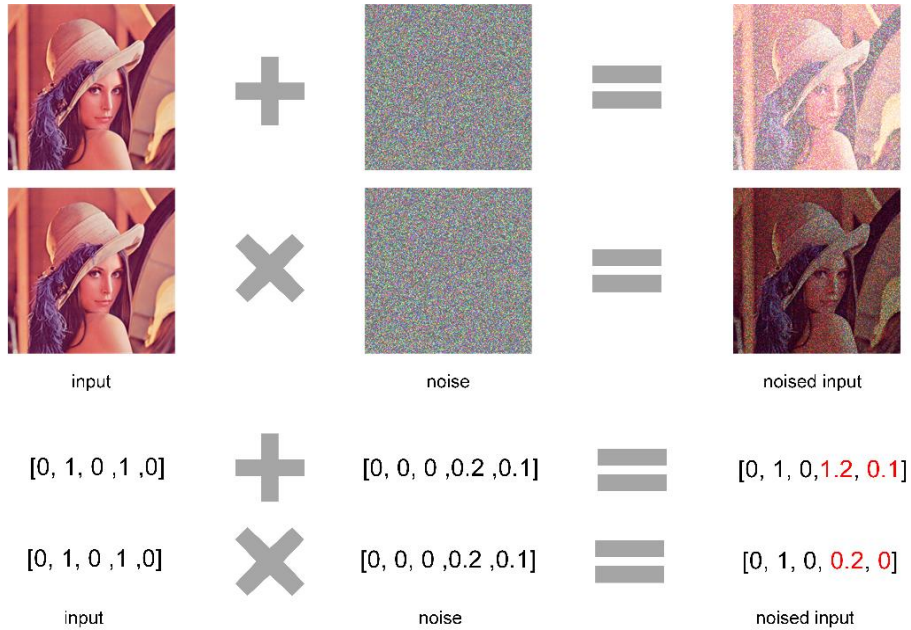

Fig. 1. Examples of additive and multiplicative noising for image and categorical data.

important, yet not fully understood, phenomenon of deep learning is an ability to learn hierarchical representations of the data. It means that deep networks "learn" complicated concepts by decomposing them into simpler ones, which are decomposed into even simpler ones etc. In case of neural networks, the deeper layer is, the more complicated concepts it can learn, basing on the previous layer. Such phenomenon is mostly observed in the convolutional networks for image processing, due to the fact that visual features are typically easily interpretable [9]. It is also a basis for layer-wise training of specific architectures like stacked autoencoders [10] and deep belief networks [11]

The hierarchical data decomposition is particularly interesting in the context of data augmentation process. This is due to the fact that artificial data can be generated not only from the input representation of the original data, but also from certain features of the data, assuming that there is a known transformation of the features to the input vector. It can be written as:

$$\hat{x} = d(\hat{z}) \qquad (15)$$

where $\hat{z}$ is a vector of features of data and $d$ is a function that produces input samples from them. Of course, to be used for the training, distribution of the samples should follow the data augmentation scheme. The hidden layers of the deep network learn the reverse mapping $d^{-1}(\hat{x})$. One can, therefore, influence the training process of hidden layers by generating artificial input data from artificial features using the above formula. In particular, one can influence the regularization of these layer, as explained in the next section.

## 4. Regularization with artificial data

Almost every type of data augmentation scheme utilizes some kind of randomness. Let us formalize this randomness by representing it as a vector $r$ of random values, called *noise vector*. Two common types of noise are an additive one:

$$\hat{x} = r + x \qquad (16)$$

and a multiplicative one:

$$\hat{x} = r * x \qquad (17)$$

Examples of both noise injection types are shown in the Figure 1. Let us now consider the case of additive noise injection. According to theoretical results of Bishop [12], adding noise to the input features is an equivalent to Tikhonov regularization. This means that training with such noise can be viewed as a regularized training, with the loss function equivalent to (7) and the regularization parameter equal to the noise variance. An exact form of the regularization component $R(\theta)$ depends on the loss function, though. Solutions for two common losses, namely mean squared error and binary crossentropy, were provided in [12].

In the Bishop's work, noise was applied to the input features only. The noise, however, can be injected not only directly to the input vector $x$ but also to high-level features, assuming that we know the function $d$ that transform them into inputs, as described by the equation (15). Therefore:

$$\hat{x} = d(\hat{z})$$
$$\hat{z} = r + z \quad (18)$$

A special case of the above will be the function $d$ that produces artificial cases from the output labels, i.e. $z = y$. We can therefore noise the output labels – this is an approach known as label smoothing, which also exhibits, albeit different, regularization properties [13]. Input noising and label smoothing are therefore two edge cases of the additive noise, applied to input and output features, respectively. A number of in-between cases of adding noise to the "hidden" features will result in Tikhonov-like regularization of the inner layers of deep network.

Considering the case of multiplicative noise, it is easy to notice similarities between it and Dropout method. More specifically, Dropout added to the input layer can be viewed as a special case of multiplicative noise, where the random component $r$ follows the Bernoulli distribution. Since Dropout is a special case of DropConnect, the multiplicative noise can be represented by the diagonal matrix $R$, but with elements following any distribution that keeps up the data augmentation scheme. On the other hand, adding Dropout/DropConnect to inner layers can be interpreted as a special case of noise injection. In this case, however, the noise is not injected directly into input features, but into the high-level features of the data:

$$\hat{x} = d(\hat{z})$$
$$\hat{z} = r * z \quad (19)$$

As a consequence of the above formula, by knowing a transformation $d$ that produces input data from high-level features $z$, it is possible to inject multiplicative noise into them, resulting in data with regularization properties analogous to that of Dropout applied to a single hidden layer.

## 5. Summary

In this paper, we have presented analogies between the regularization methods for deep learning and data augmentation process interpreted as a noise injection. It was shown that, by generating the input data from high-level features, it is possible to regularize hidden layers of the network by exploiting the ability of deep networks to learn hierarchical representations.

The analysis given here is theoretical, but there already are experimental results that partially confirm these observations. A case of convolutional neural networks for stenosis detection [14] have shown that pretraining the network on artificial dataset results in reduction of test error rate on real dataset, and, thus, smaller generalization gap. An improvement of test accuracy was also observed in the case of recurrent neural networks for ECG filtering, pretrained with synthetic signals [15]. A more definitive confirmation should be expected by the comparison of models trained for the same task with datasets created by injecting noise either into input features or high-level features of the real data.